\documentclass[11pt,a4paper]{article}
\usepackage{times,latexsym}
\usepackage{url}
\makeatletter
\g@addto@macro{\UrlBreaks}{\UrlOrds}
\makeatother
\usepackage[T1]{fontenc}
\usepackage[utf8]{inputenc}

\usepackage[acceptedWithA]{tacl2018v2}

\usepackage{xspace,mfirstuc,tabulary}

\newif\iftaclinstructions
\taclinstructionsfalse % AUTHORS: do NOT set this to true
\iftaclinstructions
\renewcommand{\confidential}{}
\renewcommand{\anonsubtext}{(No author info supplied here, for consistency with
TACL-submission anonymization requirements)}
\newcommand{\instr}
\fi

\iftaclpubformat % this "if" is set by the choice of options

\else

\fi

\usepackage{graphicx}
\usepackage{amsmath}
\usepackage{ruby}

\usepackage{booktabs}

\DeclareMathOperator*{\argmax}{argmax}
\DeclareMathOperator*{\encode}{enc}
\DeclareMathOperator*{\encodea}{enc_{\boldsymbol{\phi}}}
\DeclareMathOperator*{\encodeb}{enc_{\boldsymbol{\pi}}}
\DeclareMathOperator*{\decode}{dec}
\DeclareMathOperator*{\decodea}{dec_{\boldsymbol{\phi}}}
\DeclareMathOperator*{\decodeb}{dec_{\boldsymbol{\pi}}}
\DeclareMathOperator*{\softmaxout}{SoftmaxOut}
\DeclareMathOperator*{\softmaxouta}{SoftmaxOut_{\boldsymbol{\phi}}}
\DeclareMathOperator*{\softmaxoutb}{SoftmaxOut_{\boldsymbol{\pi}}}
\usepackage[normalem]{ulem}
\newcommand{\ms}[1]{\textcolor{red}{\bf\small}}
\usepackage{tikz}
\def\checkmark{\tikz\fill[scale=0.4](0,.35) -- (.25,0) -- (1,.7) -- (.25,.15) -- cycle;} 
\newcolumntype{R}{>{$}r<{$}}
\newcolumntype{L}{>{$}l<{$}}
\newcolumntype{C}{>{$}c<{$}}
\usepackage{bbm}

\title{Consistent Transcription and Translation of Speech}

\author{
 Matthias Sperber, Hendra Setiawan, Christian Gollan, \\
 {\bf Udhyakumar Nallasamy, Matthias Paulik} \\
 Apple \\
  {\sf \{sperber,hendra,cgollan,udhay,mpaulik\}@apple.com} \\
}

\date{}

\begin{document}
\maketitle
\begin{abstract}

The conventional paradigm in speech translation starts with a speech recognition step to generate transcripts, followed by a translation step with the automatic transcripts as input.
To address various shortcomings of this paradigm, recent work explores end-to-end trainable direct models that translate without transcribing.
However, transcripts can be an indispensable output in practical applications, which often display transcripts alongside the translations to users.

We make this common requirement explicit and explore the task of jointly transcribing and translating speech.
While high accuracy of transcript and translation are crucial, even highly accurate systems can suffer from inconsistencies between both outputs that degrade the user experience.
We introduce a methodology to evaluate consistency and compare several modeling approaches, including the traditional cascaded approach and end-to-end models.
We find that direct models are poorly suited to the joint transcription/translation task, but that end-to-end models that feature a coupled inference procedure are able to achieve strong consistency.
We further introduce simple techniques for directly optimizing for consistency, and analyze the resulting trade-offs between consistency, transcription accuracy, and translation accuracy.\footnote{We release human annotations of consistency under \url{https://github.com/apple/ml-transcript-translation-consistency-ratings}}

\end{abstract}

\section{Introduction}
Speech translation (ST) is the task of translating acoustic speech signals into text in a foreign language.
According to the prevalent framing of ST (e.g.\ \newcite{Ney1999}), given some input speech $\mathbf{x}$, ST seeks an optimal translation $\hat{\boldsymbol{t}}\in\mathcal{T}$, while possibly marginalizing over transcripts $\boldsymbol{s}\in\mathcal{S}$:
\begin{align}
\hat{\boldsymbol{t}}=&\argmax_{\boldsymbol{t}\in\mathcal{T}} \left\{P\left({\boldsymbol{t}}\mid \mathbf{x}\right)\right\}  \label{eq:marginal}\\
           \approx&\argmax_{\boldsymbol{t}\in\mathcal{T}} \left\{\sum_{\boldsymbol{s}\in\mathcal{S}}P_\text{MT}\left({\boldsymbol{t}}\mid{\boldsymbol{s}}\right)P_\text{ASR}\left({\boldsymbol{s}}\mid{\mathbf{x}}\right)\right\}.\nonumber
\end{align}

According to this formulation, ST models primarily focus on translation quality, while transcription receives less emphasis.
In contrast, practical ST user interfaces often display transcripts to the user alongside the translations. A typical example is a two-way conversational ST application that displays the transcript to the speaker for verification, and the translation to the conversation partner \cite{Hsiao2006}. Therefore, there is a mismatch between this practical requirement and the prevalent framing as described above.

While traditional ST models often do commit to a single automatic speech recognition (ASR) transcript that is then passed on to a machine translation (MT) component \cite{Stentiford1988,Waibel1991}, researchers have undertaken much effort to mitigate resulting error propagation issues by developing models that avoid making decisions on transcripts. Recent examples include direct models \cite{Weiss2017} that bypass transcript generation, and lattice-to-sequence models \cite{Sperber2017} that translate the ASR search space as a whole. Despite their merits, such models may not be ideal for scenarios that display both a translation and a corresponding transcript to users.

In this paper, we replace Eq.~\ref{eq:marginal} by a joint transcription/translation objective to reflect this requirement:
\begin{align}
\hat{\boldsymbol{s}},\hat{\boldsymbol{t}}=\argmax_{\boldsymbol{s}\in\mathcal{S},\boldsymbol{t}\in\mathcal{T}} \left\{P\left(\boldsymbol{s},\boldsymbol{t}\mid \mathbf{x}\right)\right\}. \label{eq:joint-framework}
\end{align}

This change in perspective has significant implications not only on model design but also on evaluation. First, besides translation accuracy, transcription accuracy becomes relevant and equally important. Second, the issue of \textit{consistency} between transcript and translation becomes essential. For example, let us consider a naive approach of transcribing and translating with two completely independent, potentially erroneous models. These independent models would expectedly produce inconsistencies, including inconsistent lexical choice caused by acoustic or linguistic ambiguity (Fig.~\ref{fig:lexical_consistency}), and inconsistent spelling of named entities (Fig.~\ref{fig:surface_consistency}). Even if output quality is high on average, such inconsistencies may considerably degrade the user experience.

\begin{figure}[tb!]
  \centering
\small
  \begin{tabular}{ll}
    Transcr. & \textit{And I could sort of \textbf{\textcolor{blue}{replay}} some stuff}\\
                  & \textit{that I was looking at earlier today.} \\
    \midrule
%    Transl. & \textit{Ich konnte irgendwelche Dinge \textbf{\textcolor{blue}{ersetzen}},}\\
%                & \textit{die ich heute vorhin schaute.} \\
    Transl. & \ruby{\textit{Ich} }{\tiny{I}} \ruby{\textit{konnte} }{\tiny{could}} \ruby{\textit{irgendwelche} }{\tiny{any}} \ruby{\textit{Dinge} }{\tiny{things}} \ruby{\textbf{\textcolor{purple}{\textit{ersetzen}}}}{\tiny{\textbf{\textcolor{purple}{replace}}}}\textit{,}\\
                & \ruby{\textit{die} }{\tiny{that}} \ruby{\textit{ich} }{\tiny{I}} \ruby{\textit{heute} }{\tiny{today}} \ruby{\textit{vorhin} }{\tiny{earlier}} \ruby{\textit{schaute}}{\tiny{viewed}}. \\
  \end{tabular}
  \caption{Example of lexical inconsistencies we encountered when generating transcript and translation independently. While the transcript correctly contains \textit{replay}, the German translation (mistakenly) chooses \textit{ersetzen} (English: \textit{replace}). The inconsistency is explained by the acoustic similarity between \textit{replay} and \textit{replace}, which is not obvious to a monolingual user.}
  \label{fig:lexical_consistency}
\end{figure}

Our contributions are threefold: First, we introduce the notion of consistency between transcripts and translations and propose methods to assess consistency quantitatively. Second, we survey and extend existing models, and develop novel training and inference schemes, under the hypothesis that both joint model training and a coupled inference procedure are desirable for our goal of accurate and consistent models. Third, we provide a comprehensive analysis, comparing accuracy and consistency for a wide variety of model types across several language pairs to determine the most suitable models for our task and analyze potential trade-offs.

\section{Evaluation Beyond Accuracy --- The Need for Consistency}
\label{sec:definition}

To better understand the desiderata of models that perform transcription \textit{and} translation, it is helpful to discuss how one should evaluate such models. A first step is to evaluate transcription accuracy and translation accuracy in isolation. For this purpose, we can employ well-established evaluation metrics such as word error rate (WER) for transcripts and BLEU \cite{Papineni2002} for translations. 
When considering scenarios in which both transcript and translation are displayed, \textit{consistency} is an essential additional requirement.\footnote{Other important ST use cases do not show both transcripts at the same time, such as multilingual movie subtitling. For such cases, consistency may be less critical.}  Let us first clarify what we mean by this term.

\textbf{Definition:} Consistency between transcript and translation is achieved if both are semantically equivalent, with a preference for a faithful translation approach \cite{Newmark1988}, meaning that stylistic, lexical, and grammatical characteristics should be transferred whenever fluency is not compromised. Importantly, consistency measures are defined over the space of both well-formed and erroneous sentence pairs. In the case of ungrammatical sentence pairs, consistency may be achieved by adhering to a literal or word-for-word translation strategy. % https://translationjournal.net/journal/41culture.htm

Consistency is only loosely related to accuracy, and can even be in opposition in some cases. For instance, when a translation error cannot be avoided, consistency is improved at the cost of transcription accuracy by placing the back-translated error in the transcript. Because accuracy and error metrics assess transcript or translation quality in isolation, these metrics cannot capture phenomena that involve the interplay between transcript and translation.

\begin{figure}[tb!]
  \centering
\small
  \begin{tabular}{ll}
%    \toprule
    Transcr. & \textit{\textbf{\textcolor{blue}{Bill Gross}} has several companies,}\\
                  & \textit{including one called \textbf{\textcolor{blue}{eSolar}} that has}\\
                     & \textit{some great solar \textbf{\textcolor{blue}{thermal technologie}}s.} \\
    \midrule
    Transl.1 &\textit{\textbf{\textcolor{blue}{Bill Gross}} hat mehrere Firmen, unter}\\
    (gold)         & \textit{anderem eine namens \textbf{\textcolor{blue}{eSolar}} die}\\
                      & \textit{großartige Solar\textbf{\textcolor{blue}{thermaltechnologie}}n}\\
                      & \textit{hat.} \\
    \midrule
    Transl.2 & \textit{\textbf{\textcolor{blue}{Bill Gross}} hat mehrere Firmen, eine}\\
    (incons.) & \textit{nennt sich \textbf{\textcolor{purple}{"Easolare"}}, die großartige} \\
                   & \textit{\textbf{\textcolor{purple}{Solarwärme-Technologie}} hat.}
%    \bottomrule
  \end{tabular}
  \caption{Illustration of surface-level consistency between English transcript and German translation. Only translation 1 spells both named entities (\textit{Bill Gross} and \textit{eSolar}) consistently, and the German translation \textit{Solarthermaltechnologie} (translation 1) is preferred over \textit{Solarwärme-Technologie} (translation 2), by itself a correct choice but less similar on the surface level. }
  \label{fig:surface_consistency}
\end{figure}

\subsection{Motivational Use Cases}
\label{sec:use-cases}
While ultimately user studies must assess to what extent consistency improves user satisfaction, our intention in this paper is to provide a universally useful notion of consistency that does not depend too much on specific use cases.
Nevertheless, our definition may be most convincing when put in the context of specific example use cases. 

\textbf{Lecture use case.} Here, a person follows a presentation or lecture-like event, presented in a foreign language, by reading transcript and translation on a screen \cite{Fugen2008}. This person may have partial knowledge of the source language, but knows only the target language sufficiently well. She, therefore, pays attention mainly to the translation outputs, but may occasionally consult the transcription output in case where the translation seems wrong. In this case, quick orientation can be critical, and inconsistencies would cause distraction and undermine trust and perceived transparency of the transcription/translation service.

\textbf{Dialog use case.} Next, consider the scenario of a dialog between two people who speak different languages. One person, the speaker, attempts to convey a message to the recipient, relying on an ST service that displays a transcript and a translation. Here, the transcript is shown to the speaker, who speaks only the source language, for purposes of verification and possibly correction. The translation is shown to the recipient, who only understands the target language, to convey the message \cite{Hsiao2006}. We can expect that if transcript and translation are error-free, the message is conveyed smoothly. However, when the transcript or translation contains errors, miscommunication occurs. To efficiently recover from such miscommunication, both parties should agree on the nature and details of the mistaken content. In other words, occurring errors are preferred to be consistent between transcript and translation.

\section{Estimating Consistency}
\label{sec:metrics}

Having argued for consistency as a desirable property, we now wish to empirically quantify the level of consistency between a particular model's transcripts and translations. To our knowledge, consistency has not been addressed in the context of  ST before, perhaps because traditional cascaded models have not been observed to suffer from inconsistencies in the outputs. Therefore, we propose several metrics for estimating transcript/translation consistency in this section. In \S\ref{sec:consistency_analysis}, we demonstrate strong agreement of these metrics with human ratings of consistency.

\subsection{Lexical Consistency}
\label{sec:lexical}
Our first metric focuses on semantic equivalency in general, and consistent lexical choice in particular, as illustrated in Fig.~\ref{fig:lexical_consistency}. To this end, we employ a simple lexical coverage model based on word-level translation probabilities. This approach might also capture some aspects of grammatical consistency by rewarding the use of comparable function words. We sum negative translation log-probabilities for each utterance: $t_{\boldsymbol{t}\rightarrow\boldsymbol{s}}{=}-\sum_{t_j\in\boldsymbol{t}}\max_{s_i\in\boldsymbol{s}}\log p\left(t_j\mid s_i\right)$. We then normalize across the test corpus $\mathcal{C}$ and average over both translation directions: $\frac{1}{2}\left(\frac{1}{n}\sum_{(\boldsymbol{s},\boldsymbol{t})\in\mathcal{C}}t_{\boldsymbol{t}\rightarrow\boldsymbol{s}}{+}\frac{1}{m}\sum_{(\boldsymbol{s},\boldsymbol{t})\in\mathcal{C}} t_{\boldsymbol{s}\rightarrow\boldsymbol{t}}\right)$, where $n$ and $m$ denote the number of translated and transcribed words in the corpus, respectively. In practice, we use \texttt{fast\_align} \cite{Dyer2013} to estimate probability tables from our training data. When a word has no translation probability assigned, including out-of-vocabulary cases, we use a simple smoothing method by assigning the lowest score found in the lexicon.

While it may seem tempting to employ a more elaborate translation model such as an encoder-decoder model, we deliberately choose this simple lexical approach. The main reason is that we need to estimate consistency for potentially erroneous transcript/translation pairs. In such cases, we found severe robustness issues when computing translation scores using a full-fledged encoder-decoder model. 

\subsection{Surface Form Consistency}
\label{sec:surface_cons}

Our consistency definition mentions a preference for a stylistic similarity between transcript and translation. One way of assessing stylistic aspects is to compare transcripts and translations at the \emph{surface level}. This is most sensible when the source and target language are related, and could help capture phenomena such as consistent spelling of named entities, or translations using words with similar surface form as found in the transcript. Fig.~\ref{fig:surface_consistency} provides an illustration.

We propose to assess surface form consistency through substring overlap. Our notion of substring overlap follows CharCut, which was proposed as a metric for reference-based MT evaluation \cite{Lardilleux2017}. Following Eq.~2 of that paper, we determine substring insertions, deletions, and shifts in the translation, when compared to the transcript, and compute $1-\frac{\text{deletions}+\text{insertions}+\text{shifts}}{|\boldsymbol{s}|+|\boldsymbol{t}|}$. Counts are aggregated and normalized at corpus level. To avoid spurious matches, we match only substrings of at least length $n$ (here:~5), compare in case-sensitive fashion, and deactivate CharCut's special treatment of longest common prefixes/suffixes.

 We note that  surface form consistency is less suited to language pairs that use different alphabets, and leave it to future work to explore alternatives, such as the assessment of cross-lingual phonetic similarity in such cases.

\subsection{Correlation of Transcription/Translation Error}
\label{sec:correlation}

This third metric bases consistency on well-established accuracy metrics or error metrics. We posit that a necessary (though not sufficient) condition for consistency is that the accuracy of the transcript should be correlated with the accuracy of the translation, where both are measured against some respective gold standard. We therefore propose to assess consistency through computing statistical correlation between utterance-level error metrics for transcript and translation. 

Specifically, for a test corpus of size $N$, we compute Kendall's $\tau$ coefficient across utterance-level error metrics. On the transcript side, we use utterance-level WER as the error metric. Because BLEU is a poor utterance-level metric, we make use of CharCut on the translation side, which has been shown to correlate well with human judgement at utterance level \cite{Lardilleux2017}. Formally, we compute:
\begin{align}
\mathrm{kendall}_\tau\left(\text{WER}^\text{clipped}_{1:N},\text{CharCut}_{1:N}\right).
\end{align}
Because CharCut is clipped above 1, we also apply clipping to utterance-level WER for stability.

\subsection{Combined Metric for Dialog Task}
\label{sec:comb_score}

The previous metrics estimate consistency in a fashion that is complementary to accuracy, such that it is possible to achieve good consistency despite poor accuracy. This allows trading off accuracy against consistency, depending on specific task requirements. Here, we explore a particular instance of such a task-specific trade-off that arises naturally through the formulation of a communication model. 
 We consider a dialog situation (\S\ref{sec:use-cases}), and assume that communication will be successful if and only if both transcript and translation do not contain significant deviations from some reference, as motivated in Fig~\ref{fig:dialog}. Conceptually, the main difference to \S\ref{sec:correlation} is that here we penalize, rather than reward, the \textit{bad/bad} situation (Fig~\ref{fig:dialog}). To estimate the probability of some generated transcript and translation allowing successful communication, given reference transcript and translation, we thus require that both the transcript and the translation are sufficiently accurate. For utterance with index $k$:
\begin{equation}
\begin{aligned}
P(&\text{succ}_k\mid \text{ref})=P\left(\boldsymbol{s}_k\text{ ok}\cap\boldsymbol{t}_k\text{ ok}\mid \text{ref}\right)\\
                           &= P\left(\boldsymbol{s}_k\text{ ok}\mid \text{ref}\right)\times P\left(\boldsymbol{t}_k\text{ ok}\mid\boldsymbol{s}_k,\text{ref}\right)\\
                           &\approx P\left(\boldsymbol{s}_k\text{ ok}\mid \text{ref}\right)\times P\left(\boldsymbol{t}_k\text{ ok}\mid\text{ref}\right)\\
%                         &\approx\mathrm{accuracy}(\boldsymbol{s}_k)\times\mathrm{accuracy}(\boldsymbol{t}_k).
%                         \approx&(1{-}\mathrm{WER}_k)\times(1{-}\mathrm{CharCut}_k).
\end{aligned}
\end{equation}
We then use utterance-level accuracy metrics as a proxy, computing $\mathrm{accuracy}(\boldsymbol{s}_k)=1{-}\mathrm{WER}^\text{clipped}_k$, $\mathrm{accuracy}(\boldsymbol{t}_k)=1{-}\mathrm{CharCut}_k$.
For a test corpus of size $N$ we compute corpus-level scores as $\frac{1}{N}\sum_{1\le k\le N} P(\text{succ}_k)$.

\begin{figure}[tb!]
  \centering
\small
  \begin{tabular}{ccc}
    $\boldsymbol{s}$ & $\boldsymbol{t}$ & Result \\
    \midrule
    Good & Good & Immediate success \\
    Bad & Bad & Speaker rephrases \\
    Bad & Good & Speaker rephrases unnecessarily \\
    Good & Bad & Resolve now or in later turn \\
    \bottomrule
  \end{tabular}
  \caption{Dialog use case. Whenever the transcript \emph{or} the translation has errors, additional effort is needed.}
  \label{fig:dialog}
\end{figure}
 
\section{Models for Transcription and Translation}

\begin{figure*}[tb]
\includegraphics[width=\textwidth]{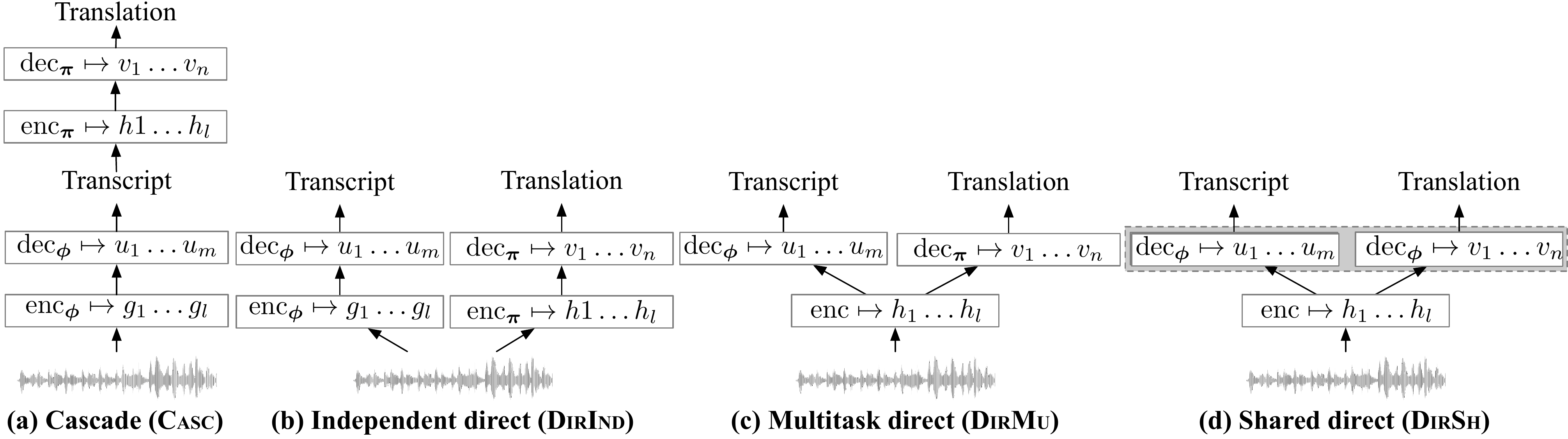}
\caption{Cascaded and direct model types.}
\label{fig:baselines}
\end{figure*}
\begin{figure}[tb]
\includegraphics[width=\columnwidth]{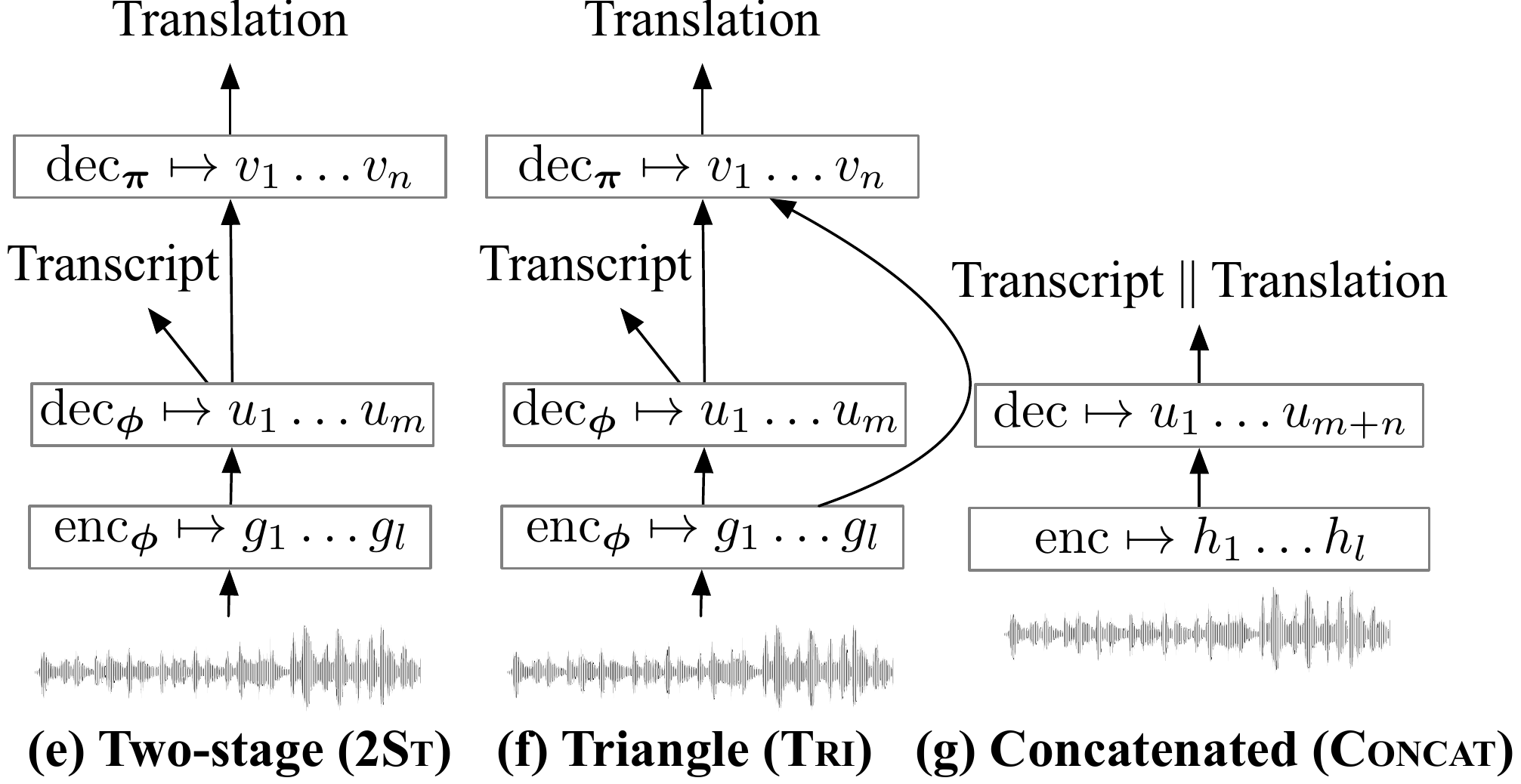}
\caption{Joint models, featuring both coupled inference and end-to-end training.}
\label{fig:joint_models}
\end{figure}

We now turn to discuss model candidates for consistent transcription and translation of speech (Figs.~\ref{fig:baselines}--\ref{fig:joint_models}). We hypothesize that there are two desirable model characteristics in our scenario. First, motivated by Eq.~\ref{eq:joint-framework}, models may achieve better consistency by performing joint inference, in the sense that no independence assumption between transcript and translation are made. We call this characteristic \textit{coupled inference}. Second, shared representations through \textit{end-to-end} (or joint) training may be of advantage in our scenario. We introduce several model variants, and also discuss whether they match these characteristics.

\subsection{Model Basics}
For a fair comparison, we keep the underlying architectural details as similar as possible across compared model types. 
All models are based on the attentional encoder-decoder framework \cite{Bahdanau2015}. 
For audio encoders, we roughly follow \newcite{Chiu2018}'s multilayer bidirectional LSTM model, which encodes log-Mel speech features that are stacked and downsampled by a factor of 3 before being consumed by the encoder. When a model requires a text encoder (\S\ref{sec:cascade}), we employ residual connections and feed-forward blocks similar to \newcite{Vaswani2017}, although for simplicity we use LSTMs \cite{Hochreiter1997} rather than self-attention in all encoder (and decoder) components. Similarly, decoder components use residual blocks of (unidirectional) LSTMs and feed-forward components \cite{Domhan2018}.

For ease of reference, we use $\encode(\cdot)$ to refer to the encoder component that transforms speech inputs (or embedded text inputs) into a hidden encoder representations, $\decode(\cdot)$ to refer to the attentional decoder component that produces hidden decoder states auto-regressively, and $\softmaxout(\cdot)$ to refer to the output softmax layer that models discrete output token probabilities. We will subscript components with the parameter sets $\boldsymbol{\pi,\phi}$ to indicate cases in which model components are separately parametrized.

\subsection{Cascaded Model (\textsc{Casc})}
\label{sec:cascade}

The cascaded model (Fig.~\ref{fig:baselines}a) represents ST's traditional approach of employing separately trained ASR and MT models \cite{Stentiford1988,Waibel1991}. Here, we employ modern sequence-to-sequence ASR and MT components. \textsc{Casc} runs a speech input $\mathbf{x}_{1:l}$ through an ASR model

\begin{equation}
\begin{aligned}
 \textbf{g}_{1:l}=&\encodea(\mathbf{x}_{1:l})\\
 \textbf{u}_i=&\decodea(\textbf{u}_{<i},\textbf{g}_{1:l},s_{i-1})\\
 P(s_i\mid \boldsymbol{s}_{<i},x_{1:l})=&\softmaxouta(\textbf{u}_i), \label{eq:cascade:asr}
\end{aligned}
\end{equation}
decodes the best hypothesis transcript $\hat{\boldsymbol{s}}$, and then applies a separate MT model
\begin{equation}
\begin{aligned}
 \textbf{h}_{1:l}=&\encodeb(\boldsymbol{\hat{s}})\\
 \textbf{v}_i=&\decodeb(\textbf{v}_{<i},\textbf{h}_{1:l},t_{i-1})\\
 P(t_i\mid \boldsymbol{t}_{<i},\boldsymbol{\hat{s}})=&\softmaxoutb(\textbf{v}_i)
\end{aligned}
\end{equation}
to generate a translation. %$\mathbf{x}_{1:l}$ are speech inputs, $\boldsymbol{s}_{1:m}$ transcripts, and $\boldsymbol{t}_{1:n}$ translations.

With respect to the two desirable characteristics of a consistent model, notice that \textsc{Casc} uses a coupled inference procedure, in the sense that no strong independence assumptions are made between transcript and translation. \textsc{Casc} may therefore be a good candidate for consistent speech transcription/translation. However, it is less straight-forward to apply end-to-end training to cascaded models.

\subsection{Direct Models}
\label{sec:direct-models}

To improve over the cascaded approach, recent work has focused on end-to-end trainable models, with direct ST models being the most prototypical end-to-end model. In the following, we describe straightforward ways of extending direct models in order to apply them to our joint transcription/translation task. Note that these direct models (Fig.~\ref{fig:baselines}b-d) generate transcripts and translations independently at inference time. In other words, these models do not support coupled inference, which may degrade consistency between transcript and translation.

It is worth discussing how our consistent transcription/translation scenario relates to the issue of error propagation, an important issue in ST in which translations are degraded due to poor transcription decisions. Prior research on direct ST models has often been motivated by the observation that direct ST models elegantly avoid the error propagation problem. However, note that by shifting perspective to the joint transcription/translation goal, error propagation loses much of its relevance. First, error propagation is usually used to describe the negative effect of intermediate decisions, but here transcripts no longer function as intermediates. Second, strategies to mitigate error propagation often seek to make translations \textit{less} influenced by transcription decisions. This is in conflict with our goal of achieving consistency between transcript and translation, which calls for precisely the opposite: transcription and translation decisions \textit{should} strongly depend on each other.

\subsubsection{Independent Direct Model (\textsc{DirInd})}
\label{sec:indp-model}

A simple way of employing direct modeling strategies for our purposes is to use two independent direct models, one for transcription, one for translation (Fig.~\ref{fig:baselines}b). Specifically, we compute

\begin{equation}
\begin{aligned}
 \textbf{g}_{1:l}=&\encodea(\mathbf{x}_{1:l})\\
 \textbf{u}_i=&\decodea(\textbf{u}_{<i},\textbf{g}_{1:l},s_{i-1})\\
 P(s_i\mid \boldsymbol{s}_{<i},x_{1:l})=&\softmaxouta(\textbf{u}_i)\\
 \textbf{h}_{1:l}=&\encodeb(\mathbf{x}_{1:l})\\
 \textbf{v}_i=&\decodeb(\textbf{v}_{<i},\textbf{h}_{1:l},t_{i-1})\\
 P(t_i\mid \boldsymbol{t}_{<i},x_{1:l})=&\softmaxoutb(\textbf{v}_i).\label{eq:direct}
\end{aligned}
\end{equation}

We are not aware of prior work using independent models for transcription and translation. We include this model as a contrastive baseline for the subsequent two models.

\subsubsection{Multitask Direct Model (\textsc{DirMu})}
\label{sec:baselines_multitask}

A major weakness of \textsc{DirInd} is that transcription and translation models are trained separately. A better solution is to follow \newcite{Weiss2017}'s approach and sharing the speech encoder between transcription and translation models while making use of multitask training. Compared to Eq.~\ref{eq:direct}, $\encodea$ and $\encodeb$ would be collapsed into a shared encoder (Fig.~\ref{fig:baselines}c). Note that originally, \newcite{Weiss2017} and follow-up works use the transcript decoder only to aid training and exploit additional data for ASR as a related task in multitask learning. However, it is straight-forward to employ the transcript decoder during inference for our purposes.

\subsubsection{Shared Direct Model (\textsc{DirSh})}
\label{sec:baselines_shared}

We can also take the amount of sharing to the extreme by sharing all weights, not just encoder weights. Increasing the amount of shared parameters may positively impact transcription/translation consistency. We are not aware of prior work using this model variant for performing speech translation. Compared to Eq.~\ref{eq:direct}, both $\encodea$/$\encodeb$ and $\decodea$/$\decodeb$ are collapsed into a shared encoder and a shared decoder (Fig.~\ref{fig:baselines}d).

\subsection{Joint Models}
\label{sec:joint_models}

We previously discussed \textsc{Casc} as a model that features coupled inference but does not support end-to-end training. We also discussed several direct models, some of which support end-to-end training, but none of which follow a coupled inference procedure. This section introduces joint models that support both end-to-end training and coupled inference.\footnote{It is worth noting that the models discussed in \S\ref{sec:joint_models} match our joint optimization goal exactly: $P(\boldsymbol{t}{\mid}\boldsymbol{s},\mathbf{x})P(\boldsymbol{s}{\mid}\mathbf{x})=P(\boldsymbol{t},\boldsymbol{s}{\mid}\mathbf{x})$. This is in contrast to \textsc{Casc} which assumes conditional independence between translation and input speech, given the transcript. However, we do not expect this to be of major importance for purposes of generating consistent transcripts and translations.}

\subsubsection{Two-Stage Model (\textsc{2St})}
\label{sec:2stage}

The two-stage model \cite{Kano2017} is conceptually close to the cascaded approach but is end-to-end trainable because continuous transcript decoder states are passed on to the translation stage. Following \newcite{Sperber2019}'s formulation, we re-use Eq.~\ref{eq:cascade:asr} to model a transcript $\boldsymbol{s}$ and hidden decoder states $\mathbf{u}_1^m$, and then compute

\begin{equation}
\begin{aligned}
 \mathbf{v}_i=&\decodeb(\mathbf{v}_{<i},\mathbf{u}_1^m)\\
 P(t_i\mid \boldsymbol{t}_{<i},\mathbf{u}_{1:m})=&\softmaxoutb(\mathbf{v}_i).
\end{aligned}
\end{equation}

Beam search is applied to decode transcripts, as well as the corresponding hidden decoder states $\mathbf{u}_{1:m}$ that are then translated. Note that in contrast to our paper, \newcite{Kano2017} and \newcite{Sperber2019} treat transcripts only as intermediate computations and do not report transcription accuracies.

\subsubsection{Triangle Model (\textsc{Tri})}
\label{sec:triangle}

The triangle model \cite{Anastasopoulos2018} extends \textsc{2St} by adding a second attention mechanism to the translation decoder that directly attends to the encoded speech inputs. Eq.~\ref{eq:cascade:asr} is reused for transcription, and translations are computed as

\begin{equation}
\begin{aligned}
& \mathbf{v}_i=\decodeb(\mathbf{v}_{<i},[\mathbf{u}_{1:m};\mathbf{h}_{1:l}],t_{i-1})\\
& P(t_i{\mid}\boldsymbol{t}_{<i},\mathbf{u}_{1:m},\mathbf{x}_{1:l}){=}\softmaxoutb(\mathbf{v}_i).
\end{aligned}
\end{equation}

\textsc{Tri} can be seen as combining \textsc{DirMu}'s advantage of featuring a direct connection between speech and translation, and \textsc{2St}'s advantage of supporting joint inference. \newcite{Anastasopoulos2018} evaluate both transcription and translation accuracy in a low-resource setting and report consistent improvements for the latter but less reliable gains for the former.

\subsubsection{Concatenated Model (\textsc{Concat})}
\label{sec:concat}

\label{sec:proposed_concatenated}

\newcite{Haghani2018} propose a sequence-to-sequence model that produces the concatenation of two outputs sequences in the context of spoken language understanding. To our knowledge it has not been employed in a ST context before, but is a very natural fit for our joint transcription/translation scenario.
\textsc{Concat} shares both the encoder and the decoder, leading to improved compactness:
\begin{equation}
\begin{aligned}
 \boldsymbol{r}_{1:m+n}:=&s_1\dots s_m t_1\dots t_n\\
 \mathbf{g}_{1:l}=&\encode(\mathbf{x}_{1:l})\\
 \mathbf{u}_i=&\decode(\mathbf{u}_{<i},\mathbf{g}_{1:l},r_{i-1})\\
 P(r_i\mid \boldsymbol{r}_{<i},\mathbf{x}_{1:l})=&\softmaxout(\mathbf{u}_i).
\end{aligned}
\end{equation}

\section{Consistency as Training and Inference Objectives}
\label{sec:cons-obj}
Having surveyed models that are suitable for our task to various degrees, we next explore simple ways to further improve the consistency of the generated outputs through adjusting training or inference objectives.

\subsection{Consistency as Training Objective}
\label{sec:optimize_train}

At training time, we wish to introduce a loss term that penalizes inconsistent outputs. While the consistency measures discussed in \S\ref{sec:metrics} are all defined at either the utterance or the corpus level, we define our loss term at the token level for convenient integration with the standard cross entropy loss term. For convenience, we opt to follow the notion of surface-level consistency (\S\ref{sec:surface_cons}), according to which we may encourage models to assign probability mass to transcript (subword) tokens that appear in the translation, and to translated tokens that appear in the transcript.\footnote{Similarly to \S\ref{sec:metrics}, this strategy targets related languages with shared alphabets, and our results for an English--German speech translation task are encouraging (\S\ref{sec:exp_directly}). We leave it to future work to explore more elaborate solutions.}

Consider the standard cross entropy loss, which is computed against the ground-truth label distribution $q(y_i)=\delta_{y_i,y^\ast_i}$ for predicted label $y_i$ at target position $i$, assigning all probability mass to the reference token $y^\ast_i$. We modify the ground truth label distribution for transcript and translation outputs, respectively:
\begin{equation}
\begin{aligned}
q_{\text{transl}}'(y_i)=(1-\epsilon)\delta_{y_i,\boldsymbol{t}_i}+\frac{\epsilon}{|\boldsymbol{s}|}\sum_{w\in\boldsymbol{s}}\delta_{y_i,w}\\
q_{\text{transcr}}'(y_i)=(1-\epsilon)\delta_{y_i,\boldsymbol{s}_i}+\frac{\epsilon}{|\boldsymbol{t}|}\sum_{w\in\boldsymbol{t}}\delta_{y_i,w}
\end{aligned}
\end{equation}

This can be seen as an instance of non-uniform label smoothing with strength $\epsilon$ \cite{Szegedy2016}. In practice, we give this loss term a relative weight of 0.1 during training, while at the same time disabling label smoothing. Because this loss requires access to the complete transcript and translation, we do not apply it at inference time.

\subsection{Consistency as Inference Objective}
\label{sec:optimize_inference}

We can also modify the inference objective to enforce more consistent outputs. A simple way for accomplishing this is via $n$-best rescoring. This is especially convenient when employing consistency measures such as lexical consistency (\S\ref{sec:lexical}), which can be computed without referring to a gold standard. Our approach here follows two simple steps: First, we compute $n$-best lists using standard beam search. Second, we select the $(\boldsymbol{s},\boldsymbol{t})$-pair that produces the best lexical consistency score. Expectedly, this rescoring approach will yield improved consistency, while possibly degrading transcript or translation accuracy. Future work may explore ways for more explicitly balancing model and consistency scores.

\section{Experimental Setup}

\subsection{Data}
We conduct experiments on the MuST-C corpus \cite{DiGangi2019c}, the largest publicly available ST corpus, containing TED\footnote{www.ted.com} talks paired with English transcripts and translations into several languages. We present results for German, Spanish, Dutch, and Russian as the target language, where the data size is 408--504 hours of English speech, corresponding to 234K--270K utterances. In TED, translated subtitles are not displayed simultaneously with the transcribed subtitles, and consistency is therefore not inherently required in this data. In practice, however, the manual translation workflow in TED results in a sufficient level of consistency between transcripts and translations. Specifically, transcripts are generated first, and translators are required to use the transcript as a starting point while also referring to the audio.\footnote{\url{www.ted.com/participate/translate}} We use MuST-C \textit{dev} for validation and report results on \textit{tst-COMMON}.

\subsection{Model and Training Details}
We make use of the 40-dimensional log Mel filterbank speech features provided with the corpus. The only text preprocessing applied to the training data is subword tokenization using \texttt{SentencePiece} \cite{Kudo2018a} with the \texttt{unigram} setting. Following most recent work on end-to-end ST models, we choose a relatively small vocabulary size of 1024, with transcription/translation vocabularies shared. No additional preprocessing steps are applied for training, but for transcript evaluation we remove punctuation and non-speech event markers such as \textit{(laughter)}, and compute case-insensitive WER. For translations, we remove non-speech markers from the decoded outputs and use \texttt{SacreBleu}\footnote{hash: case.lc+numrefs.1+smooth.4+tok.13a+version.1.4.3} \cite{Post2019} to handle tokenization and scoring.

Model hyperparameters are manually tuned for the highest accuracy with \textsc{DirMu}, our most relevant baseline. Unless otherwise noted, the same hyperparameters are used for all other model types. Weights for the speech encoder are initialized based on a pre-trained attentional ASR task that is identical to the ASR part of the direct multitask model. Other weights are initialized according to \newcite{Glorot2010}.
The speech encoder is a 5-layer bidirectional LSTM with 700 dimensions per direction. Attentional decoders consist of 2 Transformer blocks \cite{Vaswani2017} but use 1024-dimensional unidirectional LSTM instead of self-attention as a sequence model, except for the \textsc{Concat} and \textsc{DirSh} for which we increase to 3 layers. For \textsc{Casc}'s MT model, encoder/decoder both contain 6 layers with 1024-dimensional LSTMs. Subword embeddings are of size 1024.

\begin{table}[tb!]
\centering
\begin{tabular}{lcc}
Model              &  E2E training & $\boldsymbol{t}\mid \boldsymbol{s}$                          \\
\midrule 
\S\ref{sec:cascade} \textsc{Casc}    &      --          & attention         \\
\midrule 
\S\ref{sec:indp-model} \textsc{DirInd}    &       --          &    -- \\
\S\ref{sec:baselines_multitask} \textsc{DirMu}  & \checkmark &   --                \\
\S\ref{sec:baselines_shared} \textsc{DirSh}& \checkmark &  --                 \\
\midrule 
\S\ref{sec:2stage} \textsc{Tri}       & \checkmark & attention      \\
\S\ref{sec:triangle} \textsc{2St}     &  \checkmark & attention     \\
\S\ref{sec:concat} \textsc{Concat}& \checkmark & sequential    \\
\end{tabular} 
\caption{Overview of models and key properties. All models except \textsc{Casc}/\textsc{DirInd} are end-to-end (E2E) trained. Models also differ in whether translations are conditioned on transcripts ($\boldsymbol{t}{\mid}\boldsymbol{s})$, and whether conditioning is implemented through attention or through sequential decoder states.} 
\label{tab:model-overview}
\end{table}

We regularize using LSTM dropout with $p{=}0.3$, decoder input word-type dropout \cite{Gal2016}, and attention dropout, both $p{=}0.1$. We apply label smoothing with strength $\epsilon{=}0.1$. We optimize using Adam \cite{Kingma2014} with $\alpha{=}0.0005$, $\beta_1{=}0.9$, $\beta_2{=}0.98$, 4000 warm-up steps and learning rate decay by using the inverse square root of the iteration. We set the batch size dynamically based on the sentence length, such that the average batch size is 128 utterances. The training is stopped when the validation score has not improved over 3 epochs, where the validation score is the product of corpus-level translation BLEU score and corpus-level transcription word accuracy. 

For decoding and generating $n$-best lists, we use beam size 10 and polynomial length normalization with exponent $1.5$. Our implementation is based on \texttt{PyTorch} \cite{Paszke2019} and \texttt{xnmt} \cite{neubig18xnmt}, and all trainings are done using single-GPU environments, employing Tesla V100 GPUs with 32 GB memory.
 \begin{table*}[tb!]
\centering
\footnotesize
\begin{tabular}{lRRRRRRRRRRRR}

\multicolumn{1}{l}{} &
\multicolumn{3}{c}{EN$\rightarrow$DE}    &
\multicolumn{3}{c}{EN$\rightarrow$ES}    &
\multicolumn{3}{c}{EN$\rightarrow$NL}    &
\multicolumn{3}{c}{EN$\rightarrow$RU}    \\
\cmidrule(lr){2-4}
\cmidrule(lr){5-7}
\cmidrule(lr){8-10}
\cmidrule(lr){11-13}

Model & 
	\downarrow\text{WER} & \uparrow\text{BLEU} & \downarrow\text{Lex}& 
	\text{WER} & \text{BLEU} & \text{Lex}& 
	\text{WER} & \text{BLEU} & \text{Lex}& 
	\text{WER} & \text{BLEU} & \text{Lex} \\
\midrule 
SOTA \textit{cc}                       & 27.0                 & 18.5             &-                     &26.6                 &22.5               &-                      &26.6            &22.2              &-                        &27.0            &11.1                &-\\
SOTA \textit{dir}                       & -                       &  17.3            &-                    &-                       &20.8               &-                      &-                   &18.8             &-                        &-                  &8.5                  &-\\
\midrule 
\textsc{Casc}                           &   \mathbf{21.6}  &  19.3            & 10.4             & \mathbf{20.5}  & \mathbf{25.2} & 8.4             &\mathbf{20.6}&\mathbf{23.5}&10.1             &\mathbf{20.5}&13.4                & 11.3   \\ % 75, 171, 217, 216
\midrule
\textsc{DirInd}                            &  \mathbf{21.6}  & 11.0            & 21.1               & \mathbf{20.5}& 16.5                &17.8            & \mathbf{20.6}& 14.9              & 20.9            &\mathbf{20.5}&3.4                  & 29.0\\ % 81, 172, 179, 186
\textsc{DirMu}                          & 23.6                 &  18.4           & 13.9              &21.7                  &24.3                 &11.6            &23.2             &22.3              &14.3             &22.4              &13.0                 & 13.9\\ % 137;149,150, 151,
\textsc{DirSh}                       & 23.6                  & 19.0            & 14.7              & 21.3                & 24.1                 &11.5             &22.0             &22.7               &14.2             & 22.3             &13.6                & 13.6\\ % 0194, 0202, 203, 0204
\midrule 
\textsc{2St}                            & \mathit{22.2}   & \mathbf{20.1} & \mathit{9.9} &\mathit{21.4}   & 24.2               & \mathbf{7.8} & 22.6           & \mathit{23.4} & 9.4               &21.4             &\mathit{14.0} & \mathbf{10.7} \\ % 146, 174, 181, 188
\textsc{Tri}                              &  \mathit{22.2}  & \mathit{19.9} & \mathbf{9.7} &\mathit{21.0}  & \mathit{24.7} & \mathit{7.9} & 24.4           & 22.6               & \mathbf{8.9}& 21.2            &\mathbf{14.2}  & \mathbf{10.7}  \\ % 140, 173, 180, 187
\textsc{Concat}                       &  \mathit{21.9}  & 19.2              &  12.8            & \mathit{20.6}   & 23.7                &10.8             &21.9             &22.8               &   12.5           & 21.5             &13.3               & 13.3\\ % 168, 198, 199, 200
\midrule 
\end{tabular} 
\normalsize
    \caption{Comparison of WER, BLEU, lexical consistency (Lex; \S\ref{sec:lexical}) across several language pairs. We compare against state-of-the-art (SOTA) results under same data conditions by \newcite{DiGangi2019d}, where \textit{cc} denotes a cascaded model, \textit{dir} denotes a direct model. Bold font indicates the best score. Results that are not statistically significantly worse than the best score in the same column are in italics (pairwise bootstrap resampling \cite{Koehn2004}, $p{<}0.05)$.}
\label{tab:main_results}
\end{table*}

 \subsection{Human Ratings}
 \label{sec:human_ratings}
To obtain a gold standard to compare our proposed automatic consistency metrics against, we collect transcript/translation consistency ratings from human annotators. The annotators are presented a single transcript/translation pair at a time, and are asked to judge the consistency on a 4-point Likert scale. We aimed for a balanced scale which assigned a score of 4 to cases with no or only minor mismatch, a score of 3 to indicate a purely stylistic mismatch, a score of 2 to indicate a partial semantic mismatch, and a score of 1 to a complete semantic mismatch. Instructions given to the annotators include an explanation of the definition given in \S\ref{sec:definition} along with a table of several examples for each of the 4 categories. We displayed transcripts and translations in randomized order, so as to obfuscate the directionality of the translation, and do not provide the source speech utterances. Annotators are recruited from an in-house pool of trusted annotators and required to be proficient English and German speakers.

For each of the 2641 speech utterances in the MuST-C English-German test set, we collect annotations for 8 transcript/translation pairs: 7 system outputs produced by the models in Table~\ref{tab:model-overview}, and the reference transcript/translation pairs.
Each transcript/translation item is rated individually and by at least three different annotators.
In total, we employed 58 raters to produce 63412 ratings.
We fit a linear mixed-effects model on the result using the \texttt{lme4} package \cite{Bates2013}, which allows estimating the consistency of the outputs for each system, while accounting for random effects of each annotator and of each input sentence. We refer to \cite{norman2010likert,gibson2011using} for a discussion of using mixed-effects models in the context of Likert-scale ratings.

\section{Results}

We start by presenting empirical results across all four language pairs, and will then focus on English--German to discuss details. Table~\ref{tab:model-overview} contrasts the different model types that we examine.

\subsection{Accuracy Comparison}

\begin{table*}[tb!]
\centering
\footnotesize
\begin{tabular}{lrCCCRRCCC}
\multicolumn{1}{l}{} &
\multicolumn{1}{r}{}    &
\multicolumn{1}{r}{Transcript}    &
\multicolumn{2}{c}{Translation}    &
\multicolumn{5}{c}{Consistency}    \\ 
\cmidrule(lr){3-3}
\cmidrule(lr){4-5}
\cmidrule(lr){6-10}
Model                 & Params. & \downarrow\text{WER} & \uparrow\text{BLEU} & \downarrow\text{CharCut} & \downarrow\text{Lex}& \uparrow\text{Sur} & \uparrow\text{Cor} & \uparrow\text{Cmb}   & \uparrow\text{Human} \\
\midrule 
\textsc{Casc}      & $223$M & \mathbf{21.6}                & 19.2                           & 47.2                                   & 10.36                    &   10.65                   & 0.396                           &  0.474      &  3.119  \\ % 75
\midrule 
\textsc{DirInd}        &  $175$M& \mathbf{21.6}               &  11.0                          & 60.3                                   & 21.13                     & 5.24                     &  0.346                         & 0.374       &  2.195 \\ % 81
\textsc{DirMu}      & $124$M  &  23.6                           & 18.4                           &  48.7                                  & 13.89                     & 7.07                     & 0.376                          & 0.457         &  2.715  \\ % 137;
\textsc{DirSh}    & $106$M  & 23.6                            & 19.0                           &  47.9                                  & 14.71                     & 8.54                      & 0.371                         & 0.464            & 2.776  \\ % 194
\midrule 
\textsc{2St}          &  $122$M & \mathit{22.2}            & \mathbf{20.1}            & \mathbf{46.1}                     & \mathit{9.86}          &    \mathbf{12.08}    & 0.391                          & \mathbf{0.484}  &  3.170   \\ % 146
\textsc{Tri}           & $141$M & \mathit{22.2}              & \mathit{19.9}            & \mathit{46.3}                      & \mathbf{9.72}         &   \mathit{11.54}    &  \textbf{0.414}             & \mathbf{0.484}   &  \mathbf{3.192}    \\ % 140
\textsc{Concat}   &   $106$M & \mathit{21.9}             & 19.2                           & 47.1                                    & 12.79                     &   9.60                   & 0.387                          & 0.477               &  2.875    \\  % 168
\midrule 
Reference            & --                &     0                           & 100                              &  0                                      & 12.6                        & 13.3                      & 1                         &  1             & 3.594                    \\
\midrule 
\end{tabular} 
\caption{Detailed consistency results, including surface form consistency (Sur; \S\ref{sec:surface_cons}), correlation of error (Cor; \S\ref{sec:correlation}), and the combined task-specific metric (Cmb; \S\ref{sec:comb_score}). Bold font indicates the best score among automatic outputs. Results that are not statistically significantly worse than the best score in the same column are in italics.}
\label{tab:detail}
\end{table*}

To validate our implementation and to evaluate the overall model accuracy, Table~\ref{tab:main_results} compares models across four language pairs. The table confirms that, except for \textsc{DirInd}, our models obtain strong overall accuracies, as compared to prior work on the same data by \newcite{DiGangi2019d}.\footnote{Concurrent work \cite{Liu2020} obtains better transcription results, but compiles its own version of the TED corpus, thus it is unclear to what extent differences can be attributed to better data filtering strategies, which are known to be a potential issue in MuST-C.} 
 Overall, \textsc{Casc} outperforms \textsc{Concat} and the 3 direct models in terms of WER and BLEU. \textsc{2St}/\textsc{Tri} achieve similar or stronger translation accuracy compared to \textsc{Casc}. Joint model training (used by all models except \textsc{Casc} and \textsc{DirInd}) seems to hurt transcription accuracy somewhat, although the differences are often not statistically significant. This may be caused by an inherent tradeoff between translation and transcription accuracy, as discussed by \newcite{He2011}. 
 Finally, \textsc{Concat} achieves favorable transcription accuracies, while translation accuracies fall between direct models and non-direct models in most cases.

\subsection{Lexical Consistency Comparison}

Table~\ref{tab:main_results} also shows results for lexical consistency. Without exception, \textsc{2St}/\textsc{Tri} achieve the best results, followed by \textsc{Casc} and \textsc{Concat}. The direct models perform poorly in all cases. Given that \textsc{Casc} is by design a natural choice for joint transcription/translation, we did not necessarily expect \textsc{2St}/\textsc{Tri} to achieve better consistency. This encouraging evidence for the versatility of end-to-end trainable models is also supported by human ratings (\S\ref{sec:consistency_analysis}).

To categorize models regarding inference procedure and end-to-end training (Table~\ref{tab:model-overview}), we observe that coupled inference (all non-direct models) is most decisive for achieving good consistency, with conditioning on generated transcripts through sequential hidden states (\textsc{Concat}) being less effective than conditioning through attention (other non-direct models). End-to-end training also appears beneficial for consistency (\textsc{Casc} vs.\ \textsc{2St}/\textsc{Tri} and \textsc{DirInd} vs.\ \textsc{DirMu}/\textsc{DirSh}).

\subsection{Analysis of Consistency Metrics}
\label{sec:consistency_analysis}

Table~\ref{tab:detail} presents more details for English--German and includes human ratings as gold standard, along with all four proposed automatic consistency measures.
Note that the reported human ratings correspond to the intercepts of the linear mixed-effects model (\S\ref{sec:human_ratings}). The fitted model estimates the standard deviation of the random effect for annotators at 0.28 and for input sentences at 0.37. All pairwise differences between the systems in the Table are statistically significant ($p<0.01$) according to an ANOVA test.

Encouragingly, lexical and surface form consistencies are aligned, and follow the same same trends as the gold standard. The correlation-based measure agrees on the inferior consistency of direct models and the superior consistency of \textsc{Tri}, while producing slightly different orderings among the remaining models. According to our combined dialog-specific measure, \textsc{Tri}/\textsc{2St} are tied for the best overall model.

One noteworthy observation is that lexical consistency of references is far worse than for \text{2St}/\textsc{Tri} outputs. This contradicts the gold standard outputs and is possibly caused by both the system outputs and the lexical consistency score being overly literal and biased toward high-frequent outputs. For comparison against references, the surface form consistency therefore appears to be a better choice.

\subsection{Directly Optimizing for Consistency}
\label{sec:exp_directly}

Table~\ref{tab:directly} considers the English--German translation direction, and examines the effect of using strategies for direct optimization of consistency at training and inference time (\S\ref{sec:cons-obj}). All of the examined techniques improve consistency, though often at the cost of degraded accuracy. The training-time techniques appear more detrimental to transcription accuracy, while the inference-time techniques are more detrimental to translation accuracy. While \textsc{DirMu} benefits strongly from these techniques, it still falls behind \textsc{Tri}'s consistency. For \textsc{Tri}, on the other hand, surface form consistency improves to the point where it almost matches the surface form consistency between reference transcripts and translations ($3.594$, see Table~\ref{tab:detail}).

\begin{table}[tb!]
\centering
\footnotesize
\begin{tabular}{lCCRR}
Model              &\text{WER} & \text{BLEU} & \downarrow\text{Lex} & \uparrow\text{Sur}\\
\midrule 
\textsc{Tri}       & \mathbf{22.2}               & \mathbf{19.9}            &  9.72                     & 11.54  \\ % 220
training           & 24.0                              & \mathbf{19.9}           &  9.84                     & 12.09 \\ % 219
%inf/sur              & 22.3                          & 19.3                           &  9.96                    & 12.30   \\ % 221
inference              & 22.6                             & 19.5                        & \mathbf{8.79}       & \mathbf{13.17} \\% 225
%inf+train            &   24.0                            & 19.2                          & 8.90                        &  \mathbf{13.40}  \\ % 0234
\midrule 
\textsc{DirMu}    &   23.6                       & 18.4                           & 13.89                        & 7.07      \\ % 222
%inf/sur              & \mathbf{23.6}           & 18.0                          & 13.84                         & 8.03    \\ % 223
training              &  \mathbf{23.2}            &  \mathbf{18.9}          &  13.94                        &  7.94   \\ % 235
inference           & 24.0                            & 18.7                         &  \mathbf{12.63}          & \mathbf{9.29}    \\ % 224
\end{tabular} 
\caption{Direct optimization for consistency. We compare training (\S\ref{sec:optimize_train}) and inference (\S\ref{sec:optimize_inference}) approaches. Bold font indicates the best score.}
\label{tab:directly}
\end{table}

\subsection{Consistency vs.\ Accuracy}

Tables~\ref{tab:main_results}-\ref{tab:detail} tend to assign better consistency scores to models with higher accuracy scores. We wish to verify whether the trend is owed to the model characteristics or whether this indicates that our metrics fail to decouple accuracy and consistency. To this end, we again focus on English--German and introduce two new model variants: First, \textsc{CIndp} performs translation using \textsc{Casc}, but transcribes with an independently trained direct model. Expectedly, such a model shows high accuracy but low consistency, a hypothesis that is confirmed by results in Table~\ref{tab:metric}, contrasted against \textsc{DirMu}. Second, we train a weaker 2-stage model by using only half the training data. For such a model, we would expect lower accuracy but not lower consistency, which is again confirmed by Table~\ref{tab:metric}, at least to some extent (lexical consistency is worse, but the correlation measure improves). These findings indicate that accuracy and consistency are in fact reasonably well decoupled.

\begin{table}[tb!]
\centering
\footnotesize
\begin{tabular}{lRRRRRR}
\multicolumn{1}{l}{} &
\multicolumn{1}{r}{}    &
\multicolumn{1}{r}{}    &
\multicolumn{4}{c}{Consistency}    \\ 
%\cmidrule(lr){2-2}
%\cmidrule(lr){3-3}
\cmidrule(lr){4-7}

Model                                      & \text{WER}  & \text{BLEU} &  \text{Lex} & \text{Sur} & \text{Cor} & \text{Cmb}  \\
\midrule 
\textsc{DirMu}                            & 23.6            &  18.4           & 13.9       &7.1  & .38            &  .46        \\ % 
\textsc{CIndp}                          & 21.8           & 19.2            & 14.6         & 8.3  & .33            &  .47         \\ % 
\midrule 
\textsc{2St}                               & 22.2           &  20.1           & 9.9           & 12.1 & .39             & .48         \\ % 146
\textsc{2St/2}                            & 30.0           &  16.6            & 10.9        & 11.9  & .45            & .44         \\  % 206
%\midrule 

\end{tabular} 
\caption{Consistency vs.\ accuracy. \textsc{CIndp} achieves better accuracy than \textsc{DirMu}, but worse consistency scores. \textsc{2St/2} is trained on less data than \textsc{2St}, which hurts its accuracy but not its consistency scores.}
\label{tab:metric}
\end{table}

\subsection{Qualitative Analysis}
Manual inspection of the outputs of \textsc{DirMu} and \textsc{Tri} for the English--German model confirms our intuition and the quantitative findings presented above, namely that \textsc{DirMu} suffers from considerable consistency issues due to transcripts and translations being generated separately. Examples in the decoded test data are in fact easy to spot, while for \textsc{Tri} we do find any consistency problems. Figs.~\ref{fig:ex0}--\ref{fig:ex2} show cherry-picked examples.

\begin{figure}[tb]
  \centering
\footnotesize
  \begin{tabular}{lll}
    \toprule
    \textsc{DirMu} & $\boldsymbol{s}$ & \textit{\textbf{\textcolor{purple}{Doctor King}} made that shift \textbf{\textcolor{purple}{in}} thinking.} \\
                            & $\boldsymbol{t}$ & \ruby{\textit{\textbf{\textcolor{purple}{Dr. Keene }}}}{\tiny{Dr. Keene}} \ruby{\textit{machte }}{\tiny{made}} \ruby{\textit{diese }}{\tiny{this}} \ruby{\textit{Verschiebung }}{\tiny{shift}} \\
                            &                             &  \ruby{\textit{\textbf{\textcolor{purple}{und }}}}{\tiny{and}} \ruby{\textit{Denkweise.}}{\tiny{thinking.}} \\
    \midrule
    \textsc{Tri}        & $\boldsymbol{s}$ & \textit{\textbf{\textcolor{blue}{Dr. King}} made that shift \textbf{\textcolor{blue}{in}} thinking.} \\
                             & $\boldsymbol{t}$ & \ruby{\textit{\textcolor{blue}{\textbf{Dr. King }}}}{\tiny{Dr. King}} \ruby{\textit{machte }}{\tiny{made}} \ruby{\textit{diese }}{\tiny{this}} \ruby{\textit{Verschiebung }}{\tiny{shift}} \\
                              &    & \ruby{\textit{\textbf{\textcolor{blue}{im }}}}{\tiny{in}} \ruby{\textit{Denken. }}{\tiny{thinking.}} \\
    \midrule
    Ref.\                & $\boldsymbol{s}$ & \textit{See, Dr. Kean made that shift in thinking.}\\
    \bottomrule
  \end{tabular}
  \caption{Example for inconsistently spelled names and an inconsistent function word when generating transcript and translation separately using \textsc{DirMu}.}
  \label{fig:ex0}
\end{figure}

\begin{figure}[tb]
  \centering
\footnotesize
  \begin{tabular}{lll}
    \toprule
    \textsc{DirMu} & $\boldsymbol{s}$ & \textit{Really advanced civilization based on it,}\\ 
                           &                              & \textit{it \textbf{\textcolor{blue}{dances}} in in energy.} \\
                           & $\boldsymbol{t}$  & \ruby{\textit{Eine }}{\tiny A} \ruby{\textit{wirklich }}{\tiny really} \ruby{\textit{\textbf{\textcolor{blue}{fortgeschrittene }}}}{\tiny advanced} \\ 
                           &                              & \ruby{\textit{Zivilisation }}{\tiny civilization} \ruby{\textit{basiert }}{\tiny bases} \ruby{\textit{auf }}{\tiny on} \ruby{\textit{Energie.}}{\tiny energy.} \\
    \midrule
    \textsc{Tri}      & $\boldsymbol{s}$ & \textit{Really advanced civilization is based on} \\ 
                           &                              & \textit{\textbf{\textcolor{blue}{advances}} in energy.} \\
                           & $\boldsymbol{t}$   & \ruby{\textit{Wirklich }}{\tiny Really} \ruby{\textit{fortschrittliche }}{\tiny advanced} \ruby{\textit{Zivilisation }}{\tiny civilization} \\
                           &                              & \ruby{\textit{basiert }}{\tiny bases} \ruby{\textit{auf }}{\tiny on} \ruby{\textit{\textbf{\textcolor{blue}{Fortschritten }}}}{\tiny advances} \ruby{\textit{in }}{\tiny in} \ruby{\textit{Energie.}}{\tiny energy.} \\
    \midrule
    Ref.\                & $\boldsymbol{s}$ & \textit{Really advanced civilization is based on}\\
                            &                             & \textit{advances in energy.} \\
    \bottomrule
  \end{tabular}
  \caption{Here, \textsc{DirMu} makes inconsistent lexical choices for transcript and translation, leading to a correct translation despite an incorrect transcript.}
  \label{fig:ex1}
\end{figure}

\begin{figure}[tb]
  \centering
\footnotesize
  \begin{tabular}{lll}
    \toprule
    \textsc{DirMu} & $\boldsymbol{s}$ & \textit{So the question is, can you actually get}\\ 
                           &                              & \textit{that \textbf{\textcolor{blue}{desert?}}} \\
                           & $\boldsymbol{t}$ & \ruby{\textit{Die }}{\tiny The} \ruby{\textit{Frage }}{\tiny question} \ruby{\textit{ist }}{\tiny is } \ruby{\textit{also: }}{\tiny thus:} \ruby{\textit{Können }}{\tiny can} \ruby{\textit{Sie }}{\tiny you} \ruby{\textit{das }}{\tiny this} \\
                           &                             &  \ruby{\textit{tatsächlich }}{\tiny actually} \ruby{\textit{\textbf{\textcolor{blue}{verändern}}?}}{\tiny change?} \\ 
    \midrule
    \textsc{Tri} & $\boldsymbol{s}$ & \textit{So the question is: Can you actually get}\\
                      &                             &\textit{that \textbf{\textcolor{blue}{to zero?}}} \\
                      & $\boldsymbol{t}$ & \ruby{\textit{Die }}{\tiny The} \ruby{\textit{Frage }}{\tiny question} \ruby{\textit{ist }}{\tiny is} \ruby{\textit{also: }}{\tiny thus:} \ruby{\textit{Können }}{\tiny can} \ruby{\textit{Sie }}{\tiny you} \ruby{\textit{das }}{\tiny this} \\
                      &                             & \ruby{\textit{tatsächlich }}{\tiny actually} \ruby{\textit{\textbf{\textcolor{blue}{zu }}}}{\tiny to} \ruby{\textit{\textbf{\textcolor{blue}{Null }}}}{\tiny zero} \ruby{\textit{bringen?}}{\tiny bring} \\
    \midrule
    Ref. & $\boldsymbol{s}$ & \textit{So the question is: Can you actually get}\\&&\textit{that to zero?} \\
    \bottomrule
  \end{tabular}
  \caption{This is an example for \textsc{DirMu} produces incorrect outputs on both sides, with seemingly unrelated semantics.}
  \label{fig:ex2}
\end{figure}

\section{Related Work}

To our knowledge there exists no prior work on consistency for joint transcription and translation of speech in particular, or other multitask conditional sequence generation models in general. The closest related prior work is perhaps \newcite{Ribeiro2019} who analyze the case of contradictory model outputs in a question answering task in which multiple different but highly related questions are shown to the model. Other prior work examines the tradeoff between transcription and translation quality in more traditional speech translation models theoretically \cite{He2011a} and empirically \cite{He2011}. Findings indicate that optimizing for WER does not necessarily lead to the best translations in a cascaded speech translation model, which is in line with the accuracy trade-offs observed in our experiment. Concurrent work explores synchronous decoding strategies for jointly transcribing and translating speech, but does not discuss the issue of consistency \cite{Liu2020}.

With regards to our consistency evaluation metrics, a closely related line of research is work on quality estimation and cross-lingual similarity metrics \cite{Fonseca2019}. An important difference of transcription/translation consistency is that for purposes of assessing consistency there is no directionality, and both input sequences can be erroneous. It is therefore especially important for metrics to be robust against errors on both sides. Moreover, stylistic differences are often not accounted for in this line of prior work. We note the similarity of our proposed lexical consistency metric to work by \newcite{Popovic2011}, and leave it for future work to explore whether metrics from other related work can and should be employed to measure consistency.

Finally, producing transcripts alongside translations may be framed as producing an explanation (the transcript) alongside the main output (the translation). Research on explainable machine learning systems \cite[and references therein]{Smith-renner2020} may shed light on desirable properties of these explanation from a usability point of view, as well as questions related to appropriate user interface design.

\section{Conclusion}
This paper investigates the task of jointly transcribing and translating speech, which is relevant for use cases in which both transcripts and translations are displayed to users. The main theme has been the discussion of consistency between transcripts and translations. To this end, we proposed a notion of consistency and introduced techniques to estimate it. We conducted a thorough comparison across a wide range of models, both traditional and end-to-end trainable, with regards to both accuracy and consistency. As important model ingredients, we found that a coupled inference procedure, where translations are conditioned on transcripts through attention, is particularly helpful. We also found that end-to-end training improves consistency and translations but at the cost of degraded transcripts. We further introduced training and inference techniques that are effective at further improving consistency, which we found to also come with some trade-offs.

Future work should examine how consistency correlates with user experience in practice and establish specific trade-offs for various use cases. Moreover, our techniques are applicable to other multi-task use-cases that could potentially benefit from consistent outputs.

\bibliography{library}
\bibliographystyle{acl_natbib}

\end{document}